\documentclass{Interspeech2024}
\usepackage{hyperref}
\usepackage{makecell}
\usepackage{arydshln}
\interspeechcameraready 

\title{Cognitive Insights Across Languages: Enhancing Multimodal Interview Analysis}

\name[affiliation={1}]{David}{Ortiz-Perez}
\name[affiliation={1}]{Jose}{Garcia-Rodriguez}
\name[affiliation={2}]{David}{Tomás}

\address{
    $^1$ Department of Computer Technology, University of Alicante, Alicante, Spain\\
    $^2$ Department of Software and Computing Systems, University of Alicante, Alicante, Spain}
\email{dortiz@dtic.ua.es, jgarcia@dtic.ua.es, dtomas@dlsi.ua.es}


\begin{document}

\maketitle

\begin{abstract}
Cognitive decline is a natural process that occurs as individuals age. Early diagnosis of anomalous decline is crucial for initiating professional treatment that can enhance the quality of life of those affected. To address this issue, we propose a multimodal model capable of predicting Mild Cognitive Impairment and cognitive scores. The TAUKADIAL dataset is used to conduct the evaluation, which comprises audio recordings of clinical interviews. The proposed model demonstrates the ability to transcribe and differentiate between languages used in the interviews. Subsequently, the model extracts audio and text features, combining them into a multimodal architecture to achieve robust and generalized results. Our approach involves in-depth research to implement various features obtained from the proposed modalities.
\end{abstract}

\section{Introduction}\label{sec:intro}

Cognitive abilities tend to decline over time. While some abilities tend to remain unaffected, others, such as processing speed, reasoning, and memory, often show signs of deterioration \cite{age}. Although cognitive decline is a common phenomenon, some individuals may experience a more pronounced decrease in these functions, particularly those affected by diseases such as dementia \cite{dementia}. However, it is important to note that not every substantial decline in cognitive function indicates dementia. Some individuals may exhibit Mild Cognitive Impairment (MCI), which represents a stage between normal age-related cognitive deterioration and dementia \cite{mci, mci2}. Therefore, there are various stages and considerations within this spectrum. The decline in cognitive abilities due to aging poses a significant challenge in our society, as there is a trend towards an aging population \cite{ageing}.

The primary motivation behind our work focuses on the early detection of these cognitive impairments. Early detection is crucial as it facilitates prompt intervention by professionals in the field. Such interventions can greatly benefit patients by enhancing their quality of life and mitigating the decline over time, consequently stimulating positive changes in their cognitive behavior \cite{early}.

Given the aforementioned facts, we propose a multimodal and multilingual model designed for the early detection of cognitive impairment. The model aims to accurately transcribe and distinguish between various languages in audio conversations to predict the cognitive state of elderly subjects. It determines whether they exhibit signs of MCI or possess a normal cognitive state for their age, using both text transcriptions and audio data. We experiment over the TAUKADIAL dataset \cite{bib:LuzEtAlTAUKADIAL24}, which includes audio conversations where elderly individuals describe a set of images in both English and Chinese language. We also consider the actual image descriptions in our analysis.

In summary, we make the following contributions:

\begin{itemize}
    \item We propose a multimodal architecture capable of differentiating between Mild Cognitive Impairment (MCI) and a normal cognitive decline due to aging. The model takes actual conversations between a clinician and the elderly as input, extracting and post-processing transcription, and obtaining textual and acoustic information from these interactions.
    \item Our research involves a comprehensive exploration of various modalities approaches, and fusion strategies to effectively combine these modalities for optimal performance in this task.
    \item Furthermore, we emphasize the robustness and generalization of our model achieved through combinations of diverse modalities. Additionally, we highlight the ease of deployment, as the model solely relies on audio conversations.
\end{itemize}

The remainder of the paper is structured as follows. In Section \ref{sec:related}, we discuss the related works and datasets relevant to this task. Section \ref{sec:dataset} presents the employed dataset and a brief analysis. Section \ref{sec:methodology} elaborates on the various approaches employed. Section \ref{sec:experiments} introduces the experimentation and presents the obtained results. Finally, in Section \ref{sec:conclusions}, we discuss the conclusions drawn from this work.

\section{Related Works}\label{sec:related}

This section presents information on datasets used for predicting dementia and other cognitive impairments. It also covers research carried out on these datasets, with a specific focus on multimodality.

\subsection{Datasets}

Due to the sensitivity of this type of data, acquiring large volumes poses a significant challenge, constituting a crucial aspect for training deep learning architectures.

When predicting dementia or other forms of cognitive decline, some datasets focus on medical data, such as the OASIS~\cite{oasis} and ADNI \cite{adni} datasets, which incorporate medical information and Magnetic Resonance Imaging (MRI) scans of the subject's brains.
Datasets that provide information in video, audio, or text formats are relevant to our specific task and research. This emphasis relies on the practicality in real-life scenarios, where information can be captured easily using a standard camera for video recording.

Among these datasets are DementiaBank \cite{dementiabank}, and the ADReSS challenge \cite{address}. These collections share similarities, featuring audio and text transcriptions of interviews between clinicians and subjects.

\subsection{Works}

Recently, proposals for the automatic detection of cognitive impairments have made a breakthrough due to advancements in deep learning. Studies have explored patients' abilities to articulate and convey their thoughts, particularly in relation to cognitive impairments.

Significant contributions in this field include the works of Yuan et al. \cite{yuan2020disfluencies} and Balagopalan et al. \cite{balagopalan2020bert}. They used pre-trained transformer-based models, such as BERT \cite{bert}, over the ADReSS challenge to analyze text features.

Audio analysis is another relevant factor for disease detection. Studies by Hajjar et al. \cite{hajjar2023development} and Laguarta et al. \cite{audio_biomarkers} have leveraged audio biomarkers to predict cognitive diseases, such as dementia. Integration of these modalities in multimodal approaches can be achieved through various strategies, as demonstrated in our previous works \cite{ortiz2023deep} and those proposed by Sarawgi et al \cite{sarawgi2020multimodal}. A current hot topic is the fusion strategies of data for models that accommodate diverse data types, as exemplified by recent works such as CLIP \cite{clip} or LLaVA \cite{llava}.

\section{Dataset}\label{sec:dataset}

The TAUKADIAL dataset \cite{bib:LuzEtAlTAUKADIAL24} consists of recordings of elderly individuals prompted by clinicians to describe images. These recordings, which are provided in audio files, involve both Chinese and English speakers, with each participant asked to articulate descriptions for three distinct images. Notably, the image sets differ between English and Chinese participants. For English speakers, the assigned images are ``\emph{The Cookie Theft}''~\cite{cookie},``\emph{Cat Rescue}'' \cite{cat_picture}, and ``\emph{Coming and Going}'' \cite{Rockwell_1947}. Meanwhile, Chinese speakers describe a father caring for his baby, scenes from a traditional night market in Taiwan, and the daily activities in Taiwan's parks.

The dataset is annotated to indicate whether the subject has MCI or normal cognitive decline for their age. Furthermore, for each participant, a score in the form of a Mini-Mental Status Examination (MMSE) is provided. The MMSE is a commonly used clinical tool for assessing cognitive impairment in patients~\cite{mmse}. It comprises a series of eleven questions.

The training set comprises 222 samples from individuals with MCI and 165 control samples, including 74 MCI subjects and 55 control participants. For each participant, there are three samples corresponding to the described pictures. The dataset comprises 201 Chinese and 186 English samples. The participants are aged between 61 and 87, with an average of 72 years. Among the participants, 79 are female and 50 are male.

\section{Methodology}\label{sec:methodology}

In this section, we introduce the different proposed architectures.

\subsection{Data preprocessing}

This step serves as the initial phase in our work. As this dataset comprises original data in audio files, and our interest also lies in extracting the semantic meaning of interviews, a preprocessing step has been conducted to transform audio into text.

To this aim, we utilized the Whisper model \cite{whisper} to transcribe all interviews and distinguish between languages, specifically English and Chinese. The distinction between languages is crucial in this process, as the image descriptions provided for each language vary. Furthermore, translating from one language to another may not be an optimal choice, given the importance of detecting subtle grammatical errors and nuances in expression.

Concerning image descriptions, our research involved analyzing transcribed samples and following protocols outlined by the organization, leading to the acquisition of authentic descriptions corresponding to the images.
Simultaneously, audio files underwent preprocessing to extract features, such as audio biomarkers, encompassing elements like audio intensity, jitter, and shimmer, among others. To accomplish this, the OpenDBM \footnote{\url{https://github.com/AiCure/open_dbm}} library was employed. The features obtained from the openSMILE toolkit \cite{opensmile} were also extracted and tested but with no promising results.

\subsection{Text-based models}

The preprocessing of the data involves providing among other features, text transcriptions. These transcriptions represent the actual dialogue between the subjects and the clinician. During this interaction, the subject describes the image presented by the clinician, providing us with the subject's description of the image. Additionally, a genuine description of the image is also acquired. With these two sets of data, we propose the following architectures, which are tested later:

\textbf{Text.} The initial proposed model is the text model, which solely utilizes the descriptions provided by the subjects. Two separate frozen BERT models \cite{bert} are used to extract features based on the language information and description obtained during the preprocessing step. The use of two distinct BERT models is necessary as there is one for each language. In line with the Transformer architecture \cite{transformer}, a transformer encoder has been used. This encoder is trained to extract additional features and consists of two encoder layers, each with the same size as the BERT base model (768 dimensions and 8 heads). The embeddings obtained from the transformer encoder serve as an input for a final Multilayer Perceptron (MLP) used for prediction. The transformer layers, as well as the MLPs, are different for each language and also vary for the classification and regression tasks.

\textbf{Similarity.} The similarity model, as the text model, entails obtaining BERT embeddings from the subject's image descriptions. However, in this model, BERT embeddings are also acquired for the real image descriptions. Both sets of embeddings, derived from frozen models to ensure consistency during training, undergo cosine similarity analysis \cite{clip} to assess the similarity between the subject's description and the real image. The resulting cosine similarity matrix is subsequently fed into an MLP for prediction.

\textbf{Combination.} This model closely resembles the text model, with the distinction lying in the embeddings utilized to feed the transformer encoders. These embeddings result from combining the subject's descriptions with the actual descriptions. Through this amalgamation, achieved via summation, we ensure that the model duly considers the image being described.

\textbf{Combined similarity.} This ensemble closely resembles the previous one. However, in this instance, the combination of embeddings occurs between the subject's description and the similarity to the genuine description. Through this combination, the model takes into account the differences between the provided description and the actual one, along with the remainder descriptions.

With all these combinations, our aim is to ensure the capture of the most relevant information from the textual modality.

\subsection{Audio-based model}

In the context of audio modality, the features extracted during the preprocessing step have been employed. This involves a straightforward model, where the obtained features are fed into the MLP.

As the audio biomarkers do not significantly depend on the language used, no differentiation has been applied among languages for this model.

\subsection{Multimodal model}

In the case of the multimodal model, we combined the most effective text model with the audio model. We achieved this by using features extracted from the intermediate layers of their respective MLPs and combining them through concatenation. This allows us to integrate grammatical and semantic features from text-based models with those that may be lost in transcriptions.

\begin{figure*}[htbp]
  \centering
  \includegraphics[width=0.9\textwidth]{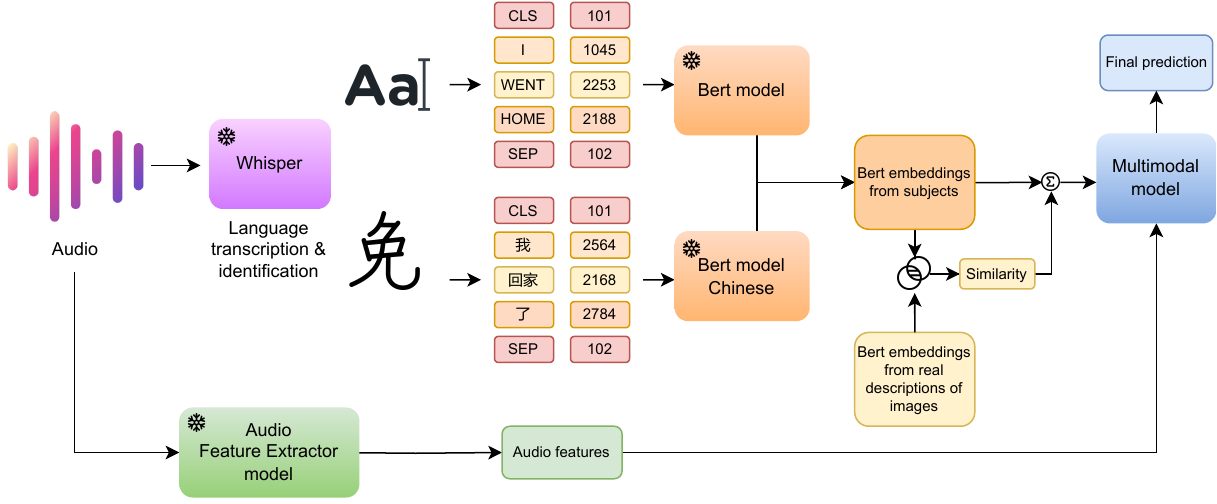}
  \caption{Overview of the multimodal architecture.}
  \label{fig:architecture}
\end{figure*}

Figure \ref{fig:architecture} illustrates the primary architecture of this pipeline, where the final MLP of the multimodal model combines features from both the audio and text models.

\section{Experiments}\label{sec:experiments}

This section introduces the experimental configurations, metrics, and results obtained from the previously proposed models.

\subsection{Experimental setup}

This subsection introduces the experimental configuration and setup used throughout this work.

\textbf{Cross-validation.}
As there is no predefined validation split, and to ensure the selection of the best and most general model, a cross-validation strategy has been employed with a fold size of 5. This entails splitting the dataset into 5 different folds, enabling the model to be evaluated on various subsets to ensure robust performance, especially with non-trainable data.

\textbf{Implementation details.}
We employed the AdamW optimizer with a learning rate of $10^{-5}$ for models based on text features and the multimodal one, and 0.01 for the audio-based ones while incorporating a weight decay of 0.01 for all of them. For the regression task, the learning rate has been increased ten times.

The models underwent training with a batch size of 16 for 100 epochs, employing an early stopping strategy with a patience of 10 epochs. The cross-entropy loss function was used for the classification task, and the mean squared error was used for the regression task.

The experiments were conducted on an NVIDIA GeForce RTX 4090. All code developed for this project is publicly available in our GitHub repository
\footnote{\url{https://github.com/davidorp/taukadial}}.

\textbf{Evaluation metrics.}
To assess the performance of our proposed models, we used the suggested metrics within the context of the TAUKADIAL challenge. Given that this challenge encompasses two distinct tasks, it employs different metrics for each. Regarding the classification task consisting of differentiating between MCI and control groups, the main metric has been the Unweighted Average Recall (UAR), expressed in Equation \ref{eq:UAR}.

\begin{equation}\label{eq:UAR}
UAR =  {\frac { \sigma + \rho }{2} }
\end{equation}

Where $\sigma$ represents specificity and $\rho$ represents sensitivity.

F1 metric is also proposed for the classification task, represented in Equation \ref{eq:F1pi}, where $\pi$ stands for the precision.

\begin{equation}\label{eq:F1pi}
F_1 =  { \frac { 2 \pi \rho}{\pi + \rho} } \textrm{,} \quad \textrm{where} \quad \pi =  { \frac { TP }{TP + FP} }
\end{equation}

For the regression tasks, the primary metric is the Root Mean Squared Error (RMSE), as shown in Equation \ref{eq:RMSE}.

\begin{equation}\label{eq:RMSE}
RMSE ={\sqrt {\frac {\sum _{i=1}^{N}({\hat {y}}_{i}-y_{i})^{2}}{N}}}
\end{equation}

The coefficient of determination $(R^2)$ metric is also proposed for the regression task, represented in Equation \ref{eq:R2}.

\begin{equation}\label{eq:R2}
R^2 =1 -
\frac {\sum_{i=1}^N(\hat{y}_{i} - y_{i})^2}
      {\sum_{i=1}^N(\hat{y}_{i} - \bar{y})^2}
\end{equation}

\subsection{Ablation study}

To evaluate the performance of different models, we conducted an ablation study using the cross-validation strategy. We implemented the proposed metrics and calculated them for each of the 5 folds with each model to compare their effectiveness.

The results for the classification task are presented in Table~\ref{tab:classification}, where values represent the mean of the 5-fold metric along with the standard deviation. We also calculated the mean and standard deviation of all metrics.

\begin{table*}[htbp]
    \begin{center}
        \caption{Results of Classification Task using 5-Fold Cross-Validation.}
        \label{tab:classification}
        \begin{tabular}{cccccc:cc}
        \hline
        \textbf{Model} & \textbf{UAR} & F1 & $\sigma$ & $\rho$ & $\pi$ & \textbf{RMSE} & $R^2$ \\
        \Xhline{1pt}
        Text  &  73.07  $\pm$  1.5  &  76.92  $\pm$  3.6  &  68.14  $\pm$  8.4  &  78.0  $\pm$  6.1  &  76.72  $\pm$  7.1  &  3.14  $\pm$  0.3  &  0.09  $\pm$  0.0 \\
        Audio  &  70.75  $\pm$  4.0  &  76.22  $\pm$  5.9  &  61.41  $\pm$  6.0  &  80.1  $\pm$  9.2  &  73.32  $\pm$  6.5 & 3.11  $\pm$  0.2  &  0.1  $\pm$  0.1 \\
        Similarity  &  68.57  $\pm$  3.3  &  72.52  $\pm$  5.1  &  63.67  $\pm$  9.8  &  73.47  $\pm$  11.8  &  73.44  $\pm$  6.0 & 2.95  $\pm$  0.4  &  \textbf{0.2  $\pm$  0.1} \\
        Combination  &  74.58  $\pm$  2.2  &  78.47  $\pm$  5.5  &  \textbf{70.44  $\pm$  12.6}  &  78.72  $\pm$  11.4  &  \textbf{79.44  $\pm$  4.3}  &  3.14  $\pm$  0.3  &  0.09  $\pm$  0.0 \\
        \begin{tabular}[c]{@{}c@{}}Combined\\similarity\end{tabular}  &  74.82  $\pm$  2.5  &  \textbf{79.43  $\pm$  1.4}  &  67.98  $\pm$  7.3  &  \textbf{81.67  $\pm$  4.3}  &  77.65  $\pm$  3.6  &  3.14  $\pm$  0.3  &  0.08  $\pm$  0.0 \\
        Multimodal  &  \textbf{75.09  $\pm$  1.8}  &  78.49  $\pm$  3.4  &  70.0  $\pm$  12.7  &  80.17  $\pm$  10.6  &  78.95  $\pm$  8.5 &  \textbf{2.93  $\pm$  0.3}  &  \textbf{0.2  $\pm$  0.1} \\
        \hline 
        \end{tabular}
    \end{center}
\end{table*}

Analyzing the obtained results from the classification task, it is evident that the multimodal model, which amalgamates features from all others, achieves the best outcome, particularly excelling in the crucial metric of UAR. Furthermore, this model also obtained the second-best result in the remaining metrics, underscoring its capability and robustness against other models. 

It is noteworthy that audio features do not offer as much information as text-based approaches, surpassing only the similarity model. However, the addition of this similarity to other features has shown improvement, as evidenced by combined models. Notably, the similarity model alone does not perform remarkably well.

Additionally, an ablation study on the regression task was conducted. Examining these results, we can conclude that the multimodal model also outperforms the others. In this case, the most important metric is the RMSE, where the multimodal model performs best. Similarly, for the coefficient of determination, both the multimodal and similarity models provide the best results. For this task, the most relevant information regarding the text differs from the other task, with the similarity to the original description being the most important part. In the case of the audio modality, it slightly improves the performance compared to the other text-based models.

\subsection{Challenge results}

As part of the TAUKADIAL challenge, a designated test set has been provided to evaluate the performance on entirely unseen data. Our submission to this challenge included four distinct models, and their respective performances are detailed in Table \ref{tab:resul}. The multimodal model was identified as the top-performing model based on the ablation study and serves as the foundation for our proposed models.

To improve the robustness of our approach, we used the multimodal model from the best fold of cross-validation as one model. Additionally, we fine-tuned this same model with data from the worst fold, resulting in the \emph{MultimodalFull} model presented in the table. This process aims to ensure that our model is exposed to the entire training dataset while maintaining a validation set.

In addition, we have presented two extra models to calculate the mean predictions for each subject, considering that each participant has three samples. This strategy ensures consistent predictions for a given participant. The models marked with an asterisk in the table represent this approach.

\begin{table}[htbp]
    \begin{center}
        \caption{Results from the challenge's test set}
        \label{tab:resul}
        \begin{tabular}{ccc}
        \hline
        \textbf{Model} & \textbf{UAR} & \textbf{RMSE} \\
        \Xhline{1pt}
        Multimodal (ours)  &  56.18  &  2.59 \\
        MultimodalFull (ours)  &  53.68  &  2.64 \\
        Multimodal* (ours)  &  55.39  &  2.58 \\
        MultimodalFull* (ours) &  52.01  &  \textbf{2.50} \\
        \hdashline
         w2v+eGeMAPs \cite{bib:LuzEtAlTAUKADIAL24} & \textbf{59.18} & 13.28 \\
         ling. \cite{bib:LuzEtAlTAUKADIAL24} & 53.59 & 2.89 \\
         w2v+ling \cite{bib:LuzEtAlTAUKADIAL24} & 50 & 4.39 \\
         eGeMAPs \cite{bib:LuzEtAlTAUKADIAL24} & 44.95 & 17 \\
         w2v \cite{bib:LuzEtAlTAUKADIAL24} & 46.05 & 4.48 \\
        \hline 
        \end{tabular}
    \end{center}
\end{table}

Examining the results obtained in the challenge. The model's performance is less effective compared to the validation folds for the classification task. Despite this, the model outperforms the results obtained in the validation set for the regression task. All models demonstrate similar performance, with the \emph{multimodal} model performing exceptionally well in classification and the \emph{Mean MultimodalFull} model standing out in the regression task.

Compared to State-Of-The-Art models, our approach achieves superior results in the regression task, surpassing others with a 2.50 RMSE score. However, in the classification task, we attain the second-best performance with a 56.18 UAR. It is important to note that our model demonstrates greater robustness and stability. It is worth mentioning that the model that outperforms ours in classification tasks exhibits significantly poorer results in regression tasks compared to our model.

\section{Conclusions}\label{sec:conclusions}

In this work, we have presented a multimodal model capable of predicting MCI or normal cognitive decline due to aging, along with a cognitive score based on MMSE. The early detection of these declines is crucial for the well-being of patients who experience them. We have employed the TAUKADIAL dataset, which consists of audio recordings from interviews between a clinician and a subject.

The research conducted throughout this work focuses on the combination of features derived from both text transcriptions and audio modalities. Consequently, we have developed a strong and easily deployable model by combining modalities.

In future works, we aspire to persist in investigating this line of research and delve into additional cognitive impairment diseases, besides their impact on people's behavior.

\section{Acknowledgements}
We would like to thank ``A way of making Europe'' European Regional Development Fund (ERDF) and MCIN/AEI/10.13039/501100011033 for supporting this work under the ``CHAN-TWIN'' project (grant TED2021-130890B- C21. HORIZON-MSCA-2021-SE-0 action number: 101086387, REMARKABLE, Rural Environmental Monitoring via ultra wide-ARea networKs And distriButed federated Learning. CIAICO/2022/132 Consolidated group project ``AI4Health'' funded by Valencian government and International Center for Aging Research ICAR funded project ``IASISTEM''. This work has also been supported by a regional grant for PhD studies, CIACIF/2022/175.

\bibliographystyle{IEEEtran}
\bibliography{mybib}

\end{document}